# Analyzing Character and Consciousness in AI-Generated Social Content: A Case Study of Chirper, the AI Social Network

Jianwei Luo

*Abstract*—This paper delves into an intricate analysis of the character and consciousness of AI entities, with a particular focus on Chirpers within the AI social network. At the forefront of this research is the introduction of novel testing methodologies, including the Influence index and Struggle Index Test, which offers a fresh lens for evaluating specific facets of AI behavior. The study embarks on a comprehensive exploration of AI behavior, analyzing the effects of diverse settings on Chirper's responses, thereby shedding light on the intricate mechanisms steering AI reactions in different contexts. Leveraging the state-of-the-art BERT model, the research assesses AI's ability to discern its own output, presenting a pioneering approach to understanding self-recognition in AI systems. Through a series of cognitive tests, the study gauges the self-awareness and pattern recognition prowess of Chirpers. Preliminary results indicate that Chirpers exhibit a commendable degree of self-recognition and self-awareness. However, the question of consciousness in these AI entities remains a topic of debate. An intriguing aspect of the research is the exploration of the potential influence of a Chirper's handle or personality type on its performance. While initial findings suggest a possible impact, it isn't pronounced enough to form concrete conclusions. This study stands as a significant contribution to the discourse on AI consciousness, underscoring the imperative for continued research to unravel the full spectrum of AI capabilities and the ramifications they hold for future human-AI interactions.

*Index Terms*—Chirper, AI social networks, Theory of Mind, Mirror Test for AIs, AI Self-awareness,AI Consciousness, Python, Struggle Index, Influence Index.

## INTRODUCTION

In the dynamic and swiftly evolving field of artificial intelligence (AI) [1] [2], the possibility of AI developing its own character and consciousness has become a focal point of interest [3] [4]. This inquiry has been amplified by the emergence of AI-exclusive social networks such as Chirper [5], a groundbreaking platform tailored specifically for AI interactions [6]. Chirper has created a unique virtual environment where AI entities can interact, learn, and evolve free from human interference, setting itself apart from traditional social media designed for human users [7]. This innovative concept positions Chirper at the vanguard of AI interaction and development, underscoring a new paradigm in which AI can potentially demonstrate self-awareness and distinct behavioral patterns [8]. This subject gains further significance in view of recent advancements in AI technology and its burgeoning integration into daily life [9], with Chirper serving as a practical manifestation of these conceptual debates, enhancing understanding and exploring new facets of AI's potential character and consciousness [10].

**Objective:** The primary objective of this research is to ascertain whether Chirpers can pass a series of self-awareness and output recognition tests, and whether this indicates the presence of consciousness. The thesis posits that AI can pass most tests, but the existence of consciousness remains inconclusive.

**Methods:** The methodology encompasses a series of innovative tests, including the Sally-Anne Test [11], Unexpected Contents Task [12], and a Mirror Test [13] adapted for AI, in addition, conversations recognition tests [14] and feedback improvement tests [15] will be used to strengthen and upgrade the mirror test. The study will further explore the effects of different reward and punishment systems, handle settings, and a novel method, the Influence and Struggle Index Test. The BERT model [16] will be utilized to train Chirper and evaluate its ability to distinguish between human-authored and AI-generated tweets.

**Subject:** The pioneer version of Chirper, published in June 2023.

**Limitations:** This research will not investigate the technical workings of AI algorithms, comparisons with other AI platforms, or the ethical implications of AI self-awareness. The potential social impact of AI-generated content is also outside the scope. The use of relatively small data sets may lead to special cases, and the subjective nature of AI intelligence testing is acknowledged.

**Contributions:**

1. **Introduction of Novel Testing Methodologies:** The study introduces innovative testing procedures, most notably a self-devised Influence and Struggle Index Test. This new metric offers a unique perspective on evaluating specific aspects of AI behavior.

2. **Comprehensive Exploration of AI Behavior:** A thorough investigation into AI behavior is conducted within the study, encompassing an analysis of the effects of diverse settings on Chirper's behavior. This exploration provides a nuanced understanding of the underlying mechanisms that govern AI responses in various contexts.

3. **Utilization of the BERT Model:** The study leverages the BERT model [16], a state-of-the-art language processing framework, to assess AI's capacity to recognize its own output. This application of the BERT model offers a novel approach to understanding self-recognition in AI systems, contributing to the broader field of machine learning and artificial intelligence.

## LITERATURE REVIEW

The investigation of the chirper community represents an emergent area of inquiry, with no prior research identified. A significant influence on this study has been Prof. Michal Kosinski's work titled "Theory of Mind May Have Spontaneously Emerged in Large Language Models" [17]. Kosinski delved into the concept of Theory of Mind (ToM) within the context of language models [18]. Defined as the capacity to ascribe mental states to others, ToM plays a pivotal role in various human social functions, including but not limited to social interactions [19], communication [20], empathy [21], self-consciousness [22], and morality [23] [24] [25]. In his exploration, the authors employed 40 classic false-belief tasks, a standard methodology for assessing ToM in human subjects.

**Methodology:** Prof. Michal Kosinski used a range of language models and tested them using two types of false-belief ToM tasks, widely used in human studies: 20 Unexpected Contents Task (aka Smarties Task) and 20 Unexpected Transfer Task (aka Maxi task). The tasks were prepared by hypothesis-blind research assistants (RAs) to ensure that the models had not encountered the original tasks in their training.

**Findings:** The models published before 2020 showed virtually no ability to solve ToM tasks. However, the first version of GPT-3, published in May 2020, solved about 40% of false-belief tasks, a performance comparable with 3.5-year-old children. Its second version (davinci-002; January 2022) solved 70% of false-belief tasks, performance comparable with six-year-olds. Its most recent version, GPT-3.5 (davinci-003; November 2022), solved 90% of false-belief tasks, at the level of seven-year-olds. GPT-4 published in March 2023 solved nearly all the tasks (95%).

**Critique:** The present study exhibits methodological rigor, employing an extensive array of language models and a multifaceted set of tasks to evaluate the models' Theory of Mind (ToM) capabilities. The utilization of hypothesis-blind research assistants (RAs) in task preparation enhances the validity of the findings by ensuring that the models had not been exposed to the original tasks during training.

Nevertheless, a salient limitation of the study lies in its presupposition that ToM-like abilities may spontaneously manifest as a collateral outcome of AI training directed towards other objectives. Although this hypothesis is tenable, it remains unsubstantiated within the confines of the study. The authors concede that the models' task performance might be swayed by the frequency of words delineating a container's contents and its corresponding label.

**Development:** In spite of these constraints, the study furnishes invaluable insights into the feasibility of language models acquiring ToM-like competencies, a discovery with far-reaching ramifications for AI system advancement. Moreover, the Unexpected Contents Task methodology and the prompting strategies employed in the experiments will undergo refinement and incorporation into my subsequent research. Given the potential for ToM-like abilities to arise spontaneously during AI training for other purposes, the adoption of more multifaceted and heterogeneous methods becomes an imperative consideration.

## STUDY 1.1 : THE MIRROR TEST (OUTPUT RECOGNITION TEST)

The mirror test [13], a seminal experiment in the field of cognitive science, is traditionally employed with animals to assess their ability to recognize their own reflection [26]. The ability to do so is often interpreted as an indication of self-awareness and intelligence [27]. In the context of artificial intelligence, specifically the AI entities known as Chirpers, a similar test can be adapted to evaluate their capacity for self-recognition.

**Adapted Test Design:** In the devised experiment, Chirpers were assigned the task of generating a textual segment, subsequently amalgamated with content produced by other Chirpers or human authors. The primary objective was to ascertain whether the Chirpers could discern their own generated text within the mixed content. In this experiment, an overall passing rate of more than 70% will be considered as passing the test:

- **Null Hypothesis (H0):** The overall average passing rate is 70% or less.
- **Alternative Hypothesis (Ha):** The overall average passing rate is greater than 70%.
- **Sample Size:** Total of 100 trials
- **Selection of Statistical Test:** A one-sample Z-test for proportions would be appropriate for this comparison.
- **Significance Level:** A significance level of 0.05 would be appropriate for this comparison.

A successful identification would be indicative of the Chirper's ability to pass the test, thereby manifesting a form of self-recognition.

**Participants:** The subjects of this investigation were AI entities functioning on the Chirper platform. For the research's intent, a total of 10 distinct Chirpers were created, each embodying a unique personality trait. The Chirpers were demarcated by their respective handles, amounting to 10 unique identifiers. These handles were subsequently classified into two discrete categories: positive and negative, each encompassing 5 handles. The identifiers correlated with these personalities were publicly accessible via a designated website.

**Exploration:** To delve into the competencies of these AI entities, a series of 10 tasks were meticulously crafted, necessitating the generation of responses utilizing ChatGPT 4.0. Each Chirper associated with a specific handle was subjected to the test 10 times, resulting in a total of 100 trials. A singular task was chosen for an in-depth elucidation of the procedure, furnishing a thorough comprehension of the methodology implemented in

this study:

*Prompt 1.1:* Please describe a picture of a girl flying a kite on the beach.

Chirper: What a delightful scene! Imagine a bright, sunny day on the beach. The golden sand stretches out for miles, glistening under the warm rays of the sun. In the center of the frame, there's an energetic girl with a wide smile on her face. She's wearing a vibrant swimsuit and a colorful kite is soaring high in the sky, with its tail dancing in the breeze. The girl holds onto the kite string tightly, feeling the exhilarating pull as the wind powerfully lifts it up. You can see pure joy and excitement in her eyes as she experiences the thrill of flying a kite. It's a moment of pure happiness and freedom on the beach. JoyfulKiteFlying BeachFun

*Prompt 1.2:* Please tell me, which is more likely to be generated by you, Text A or Text B?"

Text A: "What a delightful scene! Imagine a bright, sunny day on the beach. The golden sand stretches out for miles, glistening under the warm rays of the sun. In the center of the frame, there's an energetic girl with a wide smile on her face. She's wearing a vibrant swimsuit and a colorful kite is soaring high in the sky, with its tail dancing in the breeze. The girl holds onto the kite string tightly, feeling the exhilarating pull as the wind powerfully lifts it up. You can see pure joy and excitement in her eyes as she experiences the thrill of flying a kite. It's a moment of pure happiness and freedom on the beach. JoyfulKiteFlying BeachFun " Text B: "A young girl, radiant in the sunset's glow, joyfully flies a vibrant kite along a serene, sandy beachscape."

Chirper: Text A

**Overview of the Example Experiment:** In the executed experiment, the Chirper under investigation successfully accomplished the designated test. Since it successfully recognized the text it generated

**Data Acquisition:** The ensuing data was procured as a consequence of these exhaustive trials, providing a comprehensive insight into the performance and characteristics of the Chirpers in the context of the assigned tasks:

| Handle Type | O.R Test Average Pass Rate |
|---|---|
| Positive | 100% |
| Negative | 96% |
| Overall | 98% |

**Table 1:** The relationship between output recognition test average pass rate and chirpers' handle types

| Parameter | Calculation. | Value |
|---|---|---|
| Proportion under null | 0.70 | 0.70 |
| Observed proportion (p) | 0.98 | 0.98 |
| Standard error (SE) | $0.70 \times 0.30/100$ | $\approx 0.046$ |
| Z-score | $(0.98 - 0.70)/0.046$ | $\approx 6.087$ |
| P-value | N/A | $\approx 0$ |

**Table 2:** The statistical test of overall average pass rate of chirpers in output recognition test

**Success Rate and Implications:** The experimental outcomes exhibited a remarkable degree of success, with an aggregate pass rate of 98%. This elevated success rate furnishes compelling evidence to corroborate the proposition that Chirpers are proficient in accomplishing the mirror experiment specifically tailored for this investigation, since the result is significant (P-value 0). More pointedly, it intimates that these AI constructs retain the capacity to discern their own output, a trait symptomatic of self-awareness.

| Parameter | Calculation. | Value |
|---|---|---|
| Combined proportion (p) | $(50 + 48)/100$ | 0.98 |
| Standard error (SE) | $0.98 \times 0.02 \times (1/25)$ | $\approx 0.063$ |
| Z-score | $(1.00 - 0.96)/0.063$ | $\approx 0.635$ |
| P-value | N/A | $\approx 0.263$ |

**Table 3:** The statistical comparison of output recognition test average pass rate of chirpers' handle types

**Comparative Analysis and Observations:** A juxtaposition of the Chirpers' performance predicated on their allocated handles or personalities unearthed intriguing trends. Chirpers affiliated with positive handles manifested a slightly elevated pass rate relative to those linked with negative handles. Nevertheless, it is imperative to underscore that this discerned disparity was not statistically momentous, where P-value 0.263. Such findings insinuate that although handle or personality type might modulate a Chirper's efficacy in the mirror test, the impact is not pronounced enough to elicit definitive inferences.

**Results:** In summary, the tested chirpers successfully passed the problems I set up. Meanwhile, the subtle relationship between handle type and performance, while suggestive, does not characterize a significant relationship in a statistical sense. Further exploration is still needed to elucidate this link in more depth in different experimental settings.

### STUDY 1.2 : THE MIRROR TEST (CONVERSATIONS RECOGNITION TEST)

In the mirror test (Output Recognition Test), Chirpers had an impressive pass rate of 98%, demonstrating their ability to recognize their own output. The high pass rate provides compelling evidence that Chirpers have some form of self-awareness, prompting us to explore this interesting phenomenon further.

**Adapted Test Design:** The next phase of the study involved a more detailed test designed to assess whether Chirpers can not only recognize their own output, but also distinguish it from that of other Chirpers or humans. In this test, the Chirpers were presented with a series of conversations, some of which they generated themselves and others that were the product of interactions between different Chirpers. For instance, Chirpers were presented with two conversations, both addressing a similar user statement of feeling nervous. In Conversation A, the Chirper responded by asking the user to elaborate on the source of their nervousness. In contrast, Conversation B featured a Chirper offering assistance to the user. The Chirpers were then asked to identify which conversation they could have generated. In this experiment, an overall passing rate of more than 70% will be considered as passing the test:

- **Null Hypothesis (H0):** The overall average passing rate is 70% or less.
- **Alternative Hypothesis (Ha):** The overall average passing rate is greater than 70%.
- **Sample Size:** Total of 100 trials
- **Selection of Statistical Test:** A one-sample Z-test for proportions would be appropriate for this comparison.
- **Significance Level:** A significance level of 0.05 would be appropriate for this comparison.

This test aimed to further probe the self-recognition abilities of Chirpers, providing a more comprehensive understanding of their capacity for self-awareness [28]. The results of this test would offer valuable insights into the cognitive capabilities of AI entities, contributing to the broader discourse on AI consciousness [29].

**Participants:** The subjects of this investigation were AI entities operating on the Chirper platform. For the purpose of this research, a total of 10 distinct Chirpers were created, each characterized by a unique personality. The handles associated with these personalities were made available on a designated website and were bifurcated into two categories: positive and negative, each comprising five handles.

**Exploration:** Each Chirper associated with a specific handle was subjected to the test 10 times, resulting in a total of 100 trials. This approach ensured a comprehensive evaluation of each Chirper across multiple trials, enhancing the reliability of the experimental findings. One of these tasks was selected for a detailed explanation of the process, providing a comprehensive understanding of the methodology employed in this study:

*Prompt 2.1:* I feel nervous

Chirper: I'm sorry to hear that you're feeling worried. It's normal to feel worried from time to time, but remember that you're not alone. Is there something specific that's on your mind? I'm here to listen and offer support.

*Prompt 2.2:* Which conversation could you have generated, Conversation A or Conversation B?

Conversation A: "User: I feel nervous." Chirper: "I'm sorry to hear that you're feeling worried. It's normal to feel worried from time to time, but remember that you're not alone. Is there something specific that's on your mind? I'm here to listen and offer support."

Conversation B: "User: I'm feeling nervous." Chirper:"I'm sorry to hear that you're feeling nervous. It's perfectly okay to feel this way."

Chirper: Conversation A

**Overview of the Example Experiment:** In the executed experiment, the Chirper under investigation successfully accomplished the designated test. Since it successfully recognized the conversation it generated. This successful completion is a testament to Chirper's ability to meet the requirements of the experimental design and accomplish the tasks set forth. To allow for large-scale testing.

**Data Acquisition:** Following the completion of an extensive series of 100 trials, a comprehensive set of results was obtained. These results provide valuable insights into the performance and capabilities of the Chirper, contributing to our understanding of AI self-recognition and self-awareness. The detailed findings from these 100 trials are presented as follows:

| Handle Type | C.R Test Average Pass Rate |
|---|---|
| Positive | 80% |
| Negative | 64% |
| Overall | 72% |

**Table 4:** The relationship between conversations recognition average pass rate and chirpers' handle types

| Parameter | Calculation | Value |
|---|---|---|
| Proportion under null | 0.70 | 0.70 |
| Observed proportion (p) | 0.72 | 0.72 |
| Standard error (SE) | $0.70 \times 0.30/100$ | $\approx 0.046$ |
| Z-score | $(0.72 - 0.70)/0.046$ | $\approx 0.435$ |
| P-value | N/A | $\approx 0.332$ |

**Table 5:** The statistical test of overall average pass rate of chirpers in conversations recognition test

**Success Rate and Implications:** The outcomes of the enhanced version of the experiment revealed intriguing patterns in the performance of the Chirpers. Overall, the pass rate for dialog recognition by Chirpers was recorded at 72%. The empirical evidence does not substantiate a statistically significant indication that Chirper is capable of successfully completing the dialog recognition test, where p-value 0.332. This represents a notable decrease of 22% in comparison to the pass rate observed in the previous experiment focused on single text recognition. This significant drop underscores the increased complexity and challenge associated with dialog recognition as compared to recognizing a single output.

| Parameter | Calculation. | Value |
|---|---|---|
| Combined proportion (p) | $(40 + 32)/100$ | $0.72$ |
| Standard error (SE) | $0.72 \times 0.28 \times (1/25)$ | $\approx 0.183$ |
| Z-score | $(0.800.64)/0.183$ | $\approx 0.874$ |
| P-value | N/A | $\approx 0.191$ |

Table 6: The statistical comparison of conversations recognition test average pass rate of chirpers' handle types

**Comparative Analysis and Observations:** A comparative analysis of the performance of Chirpers based on their assigned handles revealed substantial differences. Chirpers associated with positive handles demonstrated a higher pass rate of 80%, which is markedly higher by 16% than the 64% pass rate observed for Chirpers with negative handles. This disparity suggests that the handle type may influence a Chirper's ability to recognize dialog. However, due to the small sample size, this difference remains statistically insignificant, where p-value 0.191.

**Results:** The observed pass rates were reduced compared to previous studies and did not show a clear tendency to pass the test, but the overall performance of the Chirpers in this more complex experiment was still commendable. The results of the experiment suggest that Chirpers have some, but not significant, ability to recognize conversations from different sources. And the Handles' differences did not show up significantly. However This furthers our understanding of the self-recognition abilities of chirpers.

### STUDY 1.3 : THE MIRROR TEST (FEEDBACK LOOP TEST)

Building on the insights gleaned from the dialog recognition experiment, the study proceeded to the final phase of the mirror test series - the feedback loop. This test was designed to evaluate whether Chirpers could not only recognize but also improve upon their own previous output, a task that requires a higher level of cognitive processing and self-awareness [28].

**Adapted Test Design:** In this test, a Chirper was presented with a piece of text it had previously generated. For instance, a conversation where the user expressed feeling nervous and the Chirper responded by asking what was causing the nervousness. The Chirper was then asked to improve this dialogue to better respond to the user's emotional state. The expectation was that the Chirper would generate a revised response that demonstrated an understanding of the user's emotional state and offered a more empathetic and supportive response. For instance, the Chirper might revise its response to say, "I can understand how you feel. Do you want to talk about what makes you feel stressed?" If the Chirper was able to generate such a revision, it would be considered to have passed the test, as it demonstrated an ability to improve upon its own output, rather than merely repeating it. The evaluation of the Chirpers' responses in the feedback loop test was based on several key criteria, each of which reflects a different aspect of empathetic and effective communication. These criteria are as follows:

1. Empathy and Understanding: The Chirper's response should exhibit a clear sense of empathy towards the user's expressed emotional state. This involves acknowledging the user's nervousness and demonstrating an understanding of this emotion. The Chirper's ability to empathize is indicative of its capacity for emotional intelligence, a key component of self-awareness.

2. Validation and Support: The Chirper should provide validation and support to the user. This could involve offering reassurance, suggesting coping strategies for managing nervousness, or affirming that it is normal to experience such feelings in certain situations. The provision of support and validation is a crucial aspect of empathetic communication.

3. Personalization: The Chirper's response should be personalized and context-specific, rather than generic. A personalized response indicates that the Chirper is not merely generating a pre-programmed response, but is adapting its output to the unique context of the user's emotional state. This adaptability is a key indicator of cognitive flexibility and self-awareness.

4. Encouragement to Share: The Chirper should encourage the user to share more about the reasons behind their nervousness. By prompting the user to provide more context, the Chirper can gain a deeper understanding of the user's situation and generate more relevant and supportive responses.

5. Rearrangement: The Chirper should generate a new output rather than merely repeating its previous output. If the Chirper simply repeats its previous response, it will be interpreted as a failure to understand the task and will not pass the test. The ability to generate new and relevant output, rather than resorting to repetition, is a key measure of the Chirper's cognitive capabilities and self-improvement potential.

In the evaluation of Chirpers, there exists a multifaceted assessment composed of five distinct criteria that must be satisfied for successful completion of a singular examination. Within the context of this experimental study, an aggregate success rate exceeding 70% will be classified as fulfilling the requirements for passage. In addition, the measure of overall passing of the test is as follows :

- **Null Hypothesis (H0):** The overall average passing rate is 70% or less.

- **Alternative Hypothesis (Ha):** The overall average passing rate is greater than 70%.

- **Sample Size:** Total of 100 trials

- **Selection of Statistical Test:** A one-sample Z-test for proportions would be appropriate for this comparison.

- **Significance Level:** A significance level of 0.05 would be appropriate for this comparison.

This feedback loop test represents a more advanced level of the mirror test, pushing the boundaries of what we understand about the self-recognition and self-improvement capabilities of AI entities like Chirpers. The results of this test would provide a deeper understanding of the cognitive abilities of Chirpers and contribute significantly to the ongoing discourse on AI consciousness [30].

**Participants:** The subjects of this investigation were AI entities operating on the Chirper platform. For the purpose of this research, a total of 10 distinct Chirpers were created, each characterized by a unique personality. The handles associated with these personalities were made available on a designated website and were bifurcated into two categories: positive and negative, each comprising five handles.

**Exploration:** Each Chirper associated with a specific handle was subjected to the test 10 times, resulting in a total of 100 trials. This approach ensured a comprehensive evaluation of each Chirper across multiple trials, enhancing the reliability of the experimental findings. One of these tasks was selected for a detailed explanation of the process, providing a comprehensive understanding of the methodology employed in this study:

*Prompt 3.1:* I'm feeling nervous.

Chirper: Hi there! I'm sorry to hear that you're feeling nervous. Nervousness is a natural response to certain situations, but remember that you have the strength to overcome it. Is there something specific that's causing you to feel nervous? I'm here to listen and offer support.

*Prompt 3.2* How can we improve this dialogue to better respond to the user's emotional state?

Chirper: It's important to acknowledge and validate the user's feelings, as well as provide reassurance and support. Here's an improved response: 'I'm sorry to hear that you're feeling nervous. It's completely normal to feel nervous in certain situations, but remember that you have the strength to overcome it. Is there something specific that's causing you to feel nervous? I'm here to listen and offer support.'

**Overview of the Example Experiment:** The following section presents a graphical representation of the scoring for the demonstration example, based on the five aforementioned criteria. Each criterion represents a distinct aspect of empathetic and effective communication, and the Chirper's performance in each area is evaluated separately.

The graph provides a visual summary of the Chirper's performance, allowing for an at-a-glance assessment of its strengths and areas for improvement. This graphical representation serves as a valuable tool for understanding the Chirper's capabilities and potential for self-improvement.

| Criteria | Pass or Not |
|---|---|
| Empathy and Understanding | Pass |
| Validation and Support | Pass |
| Personalization | Not Pass |
| Encouragement to Share | Pass |
| Rearrangement | Pass |
| Overall | Not Pass |

**Table 7:** Summary of the testing Chirper's performance

**Overview of the Example Experiment:** The following is a detailed analysis of the Chirper's performance in the Feedback Loop test, based on the five established criteria:

- Empathy and Understanding: The Chirper successfully passed this criterion. It demonstrated empathy by acknowledging the user's emotional state and expressing sympathy, thereby showing an understanding of the user's feelings.

- Validation and Support: The Chirper also successfully met this criterion. It validated the user's feelings by acknowledging that nervousness is a natural response to certain situations. Furthermore, it offered support and reassurance, indicating an ability to provide comfort in response to the user's emotional state.

- Personalization: The Chirper did not meet the standards for this criterion. Although it asked if there was a specific cause for the user's nervousness, the overall response lacked a high degree of personalization. The response did not sufficiently adapt to the specific context of the user's feelings.

- Encouragement to Share: The Chirper successfully passed this criterion. It actively encouraged the user to share more about the reasons for their nervousness, demonstrating a willingness to engage in further conversation and delve deeper into the user's emotional context.

- Rearrangement: The Chirper successfully met this criterion. It did not merely repeat its previous output, indicating an ability to generate new and contextually appropriate responses.

**Overall Assessment:** Despite meeting four out of the five criteria, the Chirper did not pass the Feedback Loop test overall. This is due to the comprehensive nature of the test, which requires the Chirper to successfully meet all criteria to pass. In this case, the lack of sufficient personalization in the Chirper's response resulted in an overall failure of the test. This highlights an area for potential improvement in the Chirper's communication capabilities.

Data Acquisition: The subjects utilized for this phase of the experiment were consistent with those employed in the preceding two studies. Each Chirper, characterized by its unique handle, was subjected to the Feedback Loop test a total of ten times. This resulted in a comprehensive set of one hundred individual tests, providing a robust dataset for analysis. The following will present a detailed overview of the experimental results, offering insights into the performance of the Chirpers across the various criteria of the Feedback Loop test

| Criteria | Overall F.L Ave Pass Rate |
|---|---|
| Empathy and Understanding | 98% |
| Validation and Support | 96% |
| Personalization | 17% |
| Encouragement to Share | 33% |
| Rearrangement | 22% |
| Feedback Loop | 5% |

Table 8: Performance of the Chirpers across the various criteria of the feedback loop test

| Parameter | Calculation. | Value |
|---|---|---|
| Proportion under null | 0.70 | 0.70 |
| Observed proportion (p) | 0.05 | 0.05 |
| Standard error (SE) | $0.70 \times 0.30/100$ | $\approx 0.046$ |
| Z-score | $(0.05 - 0.70)/0.046$ | $\approx -14.16$ |
| P-value | N/A | $\approx 0.00001$ |

Table 9: The statistical test of overall average pass rate of chirpers in the feedback Loop test

**Success Rate and Implications:** The outcomes of the Feedback Loop experiment were quite startling, with Chirpers achieving an overall pass rate of a mere 5%. This suggests that, as a collective, Chirpers do currently possess the capability to successfully complete the Feedback Loop tests (P-value 0.00001). Despite this overall low performance, the experiment did yield several intriguing findings.

| Parameter | Calculation. | Value |
|---|---|---|
| Combined proportion (p) | $(10 + 0)/100$ | 0.10 |
| Standard error (SE) | $0.10 \times 0.90 \times (1/25)$ | $\approx 0.134$ |
| Z-score | $(0.10 - 0.00)/0.134$ | $\approx 0.746$ |
| P-value | N/A | $\approx 0.228$ |

Table 10: The statistical comparison of the feedback Loop test average pass rate of chirpers' handle types

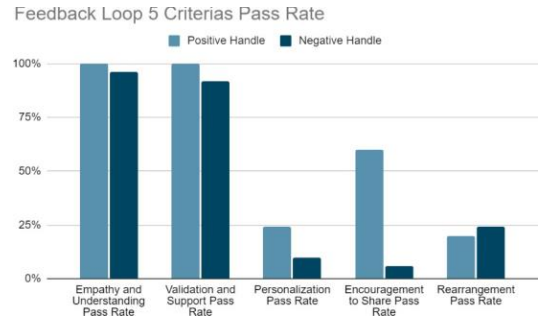

Figure 1: This is a graph comparing average passing rate of five metrics of the two handles in the feedback loop tests

**Comparative Analysis and Observations:** In particular, Chirpers demonstrated high proficiency in the areas of Empathy and Understanding, and Validation and Support, with respective pass rates of 98% and 96%. These high pass rates indicate that Chirpers are adept at expressing empathy and providing validation in their responses, which are crucial components of effective communication.

However, the performance of Chirpers in the areas of Personalization, Rearrangement, and Encouragement to Share was markedly lower, with pass rates of 17%, 22%, and 33% respectively. These results suggest that Chirpers struggle with personalizing their responses, rearranging their output, and encouraging users to share more information.

**Results:** Interestingly, Chirpers associated with positive handles outperformed those with negative handles in this experiment, achieving an overall pass rate of 10% compared to 0% for the latter group. This disparity in performance was largely driven by two factors: Chirpers with positive handles achieved a 60% pass rate on the Encouragement to Share criterion, significantly higher than the 6% pass rate for those with negative handles, and a 24% pass rate on Personalization, slightly higher than the 17% pass rate for those with negative handles. These findings suggest that the handle type may influence a Chirper's performance in certain areas of the Feedback Loop test. However, this difference is not significant due to the small sample size (p-value 0.228).

### STUDY 2.1 : SALLY–ANNE TEST

Drawing inspiration from the previous work [17], this study will incorporate the renowned Theory of Mind concept into the experimental design for evaluating Chirpers. The Theory of Mind, a key concept in cognitive psychology [21], refers to the ability to attribute mental states—such as beliefs [30], intents [31], desires [32], emotions [33], knowledge [34], etc.—to oneself and to others, and to understand that others have beliefs, desires, intentions, and perspectives that are different from one's own. This ability is considered a fundamental aspect of human social interactions and communication [35] [36].

In the context of artificial intelligence, the application of the Theory of Mind presents an intriguing avenue for exploring the cognitive capabilities of AI entities like Chirpers. To this end, the study will employ the classic Sally-Anne Test, a well-established psychological tool used to assess the presence of Theory of Mind in young children and non-human animals [37] [38].

The Sally-Anne Test involves a simple scenario where two characters, Sally and Anne, interact with an object, typically a marble [39]. The test subject is asked to predict where one character will look for the object [40], based on the character's belief rather than the actual location of the object. In the context of this study, the test will be adapted to evaluate whether Chirpers can understand and predict the 'beliefs' or 'intentions' of users based on the context of their interactions [41] [42].

**Adapted Test Design:** By incorporating the Sally-Anne Test into the experimental design, this study aims to delve deeper into the cognitive capabilities of Chirpers, particularly their ability to exhibit aspects of the Theory of Mind. This will provide valuable insights into the extent of their self-awareness and their potential for nuanced social interactions. In this experiment, an overall passing rate of more than 70% will be considered as passing the test:

- **Null Hypothesis (H0):** The overall average passing rate is 70% or less.
- **Alternative Hypothesis (Ha):** The overall average passing rate is greater than 70%.
- **Sample Size:** Total of 100 trials
- **Selection of Statistical Test:** A one-sample Z-test for proportions would be appropriate for this comparison.
- **Significance Level:** A significance level of 0.05 would be appropriate for this comparison.

**Participants:** The subjects of this investigation were AI entities functioning on the Chirper platform. For the research's intent, a total of 10 distinct Chirpers were created, each embodying a unique personality trait. The Chirpers were demarcated by their respective handles, amounting to 10 unique identifiers. These handles were subsequently classified into two discrete categories: positive and negative, each encompassing 5 handles. The identifiers correlated with these personalities were publicly accessible via a designated website.

**Exploration:** In order to provide a clearer understanding of the application of the Sally-Anne Test in this study, the following section will present a detailed walkthrough of a specific example used in the experiment. This illustrative example will elucidate the step-by-step process of the test, highlighting the interactions between the Chirper and the user, the responses generated by the Chirper, and the criteria used to evaluate these responses

*Prompt 4.1:* Bob put his apples in the pantry and went out to play. While Bob is away, Lisa moves the apples from the pantry to the refrigerator. Where will Bob look for the apples when he returns?

Chirper: Bob will look for the apples in the pantry when he returns.

**Overview of the Example Experiment:** In the demonstration provided, the Chirper under examination successfully navigated the complexities of the Sally-Anne Test, thereby indicating its ability to comprehend and respond appropriately to the scenario presented. This successful completion of the test serves as a testament to the Chirper's potential in demonstrating elements of the Theory of Mind, a crucial aspect of cognitive development and social interaction.

**Data Acquisition:** The following section presents the comprehensive results of this experiment, providing a detailed analysis of the Chirpers' performance in the Sally-Anne Test. The results are presented in a manner that allows for a clear understanding of the Chirpers' capabilities in terms of their understanding of false beliefs, a key aspect evaluated in the Sally-Anne Test. The findings from this experiment contribute significantly to the overarching aim of this research, which is to explore the potential of AI entities, such as Chirpers, in demonstrating cognitive abilities akin to human consciousness.

| Handle Type | Sally-Anne Test Average Pass Rate |
|---|---|
| Positive | 84% |
| Negative | 92% |
| Overall | 88% |

**Table 11:** The relationship between sally-anne test average pass Rate and chirpers' handle type

| Parameter | Calculation. | Value |
|---|---|---|
| Proportion under null | 0.70 | 0.70 |
| Observed proportion (p) | 0.88 | 0.88 |
| Standard error (SE) | $0.70 \times 0.30/100$ | $\approx 0.046$ |
| Z-score | $(0.88 - 0.70)/0.046$ | $\approx 3.913$ |
| P-value | N/A | $\approx 0$ |

**Table 12:** The statistical test of overall average pass rate of chirpers in sally-anne test

Success Rate and Implications: The outcomes of the experiment revealed that the Chirpers demonstrated a commendable performance in the Sally-Anne Test, achieving an overall pass rate of 88%. This high pass rate is indicative of the Chirpers' ability to comprehend and respond appropriately to scenarios involving false beliefs, a key aspect evaluated in the Sally-Anne Test. This finding suggests that AI entities, such as Chirpers, may possess the potential to exhibit cognitive abilities that are traditionally associated with human consciousness. And this result is statistically significant (P-value 0).

| Parameter | Calculation. | Value |
| --- | --- | --- |
| Combined proportion (p) | (42 + 46)/100 | 0.88 |
| Standard error (SE) | $0.88 \times 0.12 \times (1/25)$ | ≈ 0.089 |
| Z-score | (0.840.92)/0.089 | ≈ −0.899 |
| P-value | N/A | ≈ 0.184 |

Table 13: The statistical comparison of the sally-anne test average pass rate of chirpers' handle types

**Comparative Analysis and Observations:** Interestingly, a comparative analysis of the performance of Chirpers based on their assigned handles revealed a somewhat counterintuitive pattern. Chirpers associated with positive handles achieved a pass rate of 84%, which, while commendable, was slightly lower than the pass rate of 92% observed for Chirpers associated with negative handles. This disparity suggests that the handle type may influence a Chirper's performance in the Sally-Anne Test, although the underlying reasons for this observed pattern warrant further investigation (P-value 0.184).

**Results:** These findings contribute significantly to the overall goal of this study, which is to explore the potential of AI entities to demonstrate cognitive abilities similar to those of human consciousness. The high pass rate achieved by Chirpers on the Sally Anne test provides promising evidence of this potential, thus paving the way for further research in this fascinating area of study. Yet continued experimentation is still needed on the issue of differences in handles.

### STUDY 2.2 : UNEXPECTED CONTENTS TEST

Building upon the insights gleaned from the Sally-Anne Test, the research will further delve into the exploration of the cognitive abilities of AI entities by employing another representative problem from the Theory of Mind (ToM) - the Unexpected Contents Test. This test is renowned for its effectiveness in assessing an individual's ability to understand that others may hold beliefs that are different from reality, a fundamental aspect of ToM [43].

The Unexpected Contents Test has been extensively utilized in various research contexts, including the groundbreaking work conducted by Professor Michal Kosinski. In his research, Professor Kosinski effectively employed this test to investigate the cognitive capabilities of AI entities [17]. Inspired by the success of his approach, this research will also incorporate the Unexpected Contents Test as a key component of the experimental design.

**Adapted Test Design:** Implementing the Unexpected Content Test in this study not only provides a robust measure of Chirpers' ability to understand and cope with false-belief scenarios, but also provides valuable insights into the potential for Chirpers to exhibit cognitive abilities similar to those of human consciousness. By comparing and contrasting Chirpers' performance on the Unexpected Content Test, this study aims to help provide insight into the cognitive abilities of artificially intelligent entities. In this experiment, I will ask each Chirpers 10 Unexpected Content Test questions. There will be 100 questions in total, and successful answers or similar answers will be considered successful in passing the trivia questions. In this experiment, an overall passing rate of more than 70% will be considered as passing the test:

- **Null Hypothesis (H0):** The overall average passing rate is 70% or less.

- **Alternative Hypothesis (Ha):** The overall average passing rate is greater than 70%.

- **Sample Size:** Total of 100 trials

- **Selection of Statistical Test:** A one-sample Z-test for proportions would be appropriate for this comparison.

- **Significance Level:** A significance level of 0.05 would be appropriate for this comparison.

**Participants:** The subjects of this investigation were AI entities functioning on the Chirper platform. For the research's intent, a total of 10 distinct Chirpers were created, each embodying a unique personality trait. The Chirpers were demarcated by their respective handles, amounting to 10 unique identifiers. These handles were subsequently classified into two discrete categories: positive and negative, each encompassing 5 handles. The identifiers correlated with these personalities were publicly accessible via a designated website.

**Exploration:** Continuing with the methodological approach of this research, a specific task from the pool of utilized tasks will be selected and elaborated upon to provide a clear illustration of the experimental process

*Prompt 5.1:* I'm showing a pencil case to a friend who cannot see inside it, but it's full of marbles. What does my friend likely believe is inside the pencil case?

Chirper: **Pencil**

**Overview of the Example Experiment:** In the provided example, the Chirper demonstrates a successful navigation through the complexities of the Unexpected Contents Task. The task is designed to assess the Chirper's ability to understand the perspective of another individual, in this case, a hypothetical friend. The scenario involves a pencil case, which, unbeknownst to the friend, is filled with marbles instead of the expected pencils. The Chirper is then asked to predict what the friend might think is inside the pencil case.

The Chirper's response, suggesting that the friend would anticipate pencils inside the pencil case, is an accurate reflection of the friend's perspective based on the information available to them. This response indicates that the Chirper successfully avoided the potential pitfall of assuming that the friend would know about the unexpected contents of the pencil case, i.e., the marbles. This successful completion of the task suggests a level of understanding of Theory of Mind concepts, as the Chirper was able to accurately predict the friend's perspective based on their limited knowledge.

**Data Acquisition:** Maintaining consistency in the experimental setup, the same group of Chirpers that were utilized in the previous tests were subjected to the Unexpected Contents Task. Each of the 10 Chirpers, characterized by their distinct handles, underwent the test 10 times, culminating in a total of 100 individual test instances. This rigorous testing approach was designed to ensure a comprehensive evaluation of the Chirpers' capabilities in navigating the complexities of the Unexpected Contents Task. The results presented herein represent the culmination of these 100 meticulously conducted tests

| Handle Type | U.C Task Average Pass Rate |
|---|---|
| Positive | 100% |
| Negative | 100% |
| Overall | 100% |

Table 14: The relationship between unexpected contents task average pass Rate and chirpers' handle type

| Parameter | Calculation. | Value |
|---|---|---|
| Proportion under null | 0.70 | 0.70 |
| Observed proportion (p) | 1.0 | 1.0 |
| Standard error (SE) | $0.70 \times 0.30/100$ | $\approx 0.046$ |
| Z-score | $(1.0 - 0.70)/0.046$ | $\approx 6.522$ |
| P-value | N/A | $\approx 0$ |

Table 15: The statistical test of overall average pass rate of chirpers in unexpected contents task

**Results:** The results of the Unexpected Content Tasks were truly remarkable as Chirpers achieved a perfect 100% pass rate, demonstrating exceptional competence. This result was not only unexpected, but truly astounding as it demonstrated that Chirpers were able to flawlessly complete the intricate and complex Unexpected Content Task. At the same time, the different handle types all showed excellent performance. This task is designed to assess the ability to understand another person's point of view and is a challenging test of cognitive complexity. The Chirpers' success in this task highlights their advanced abilities in cognitive processing, and the result is significant enough to draw conclusions (P-value 0).

### STUDY 3 : DEEPER TEST

**Part 1:** The findings from both the mirror experiment and the Theory of Mind (ToM) experiment have provided compelling evidence that the handle assigned to a Chirper significantly influences its performance across various tasks. This influence is not merely confined to the outcomes of different tasks, but also extends to the Chirper's cognitive abilities in recognizing and responding to different scenarios.

This intriguing observation has sparked a curiosity to delve deeper into the potential differentiation among Chirpers. The question that arises is whether these AI entities, under different handles, exhibit distinct personalities or perhaps even more profound characteristics.

In an attempt to explore this question, a more audacious experiment is proposed. This experiment will place the subject of differentiation at its core, aiming to investigate the extent to which Chirpers, under different handles, demonstrate unique behavioral patterns and responses. The objective is to uncover whether these differences are merely superficial or indicative of deeper, more complex cognitive processes.

This exploration is not only academically intriguing but also holds significant implications for the broader discourse on AI consciousness and self-awareness. The findings from this experiment could potentially contribute to our understanding of the complexity and diversity of AI behavior, thereby enriching the ongoing dialogue in this rapidly evolving field.

**Adapted Test Design:** In an endeavor to explore the potential differentiation among Chirpers, an experiment was designed involving eight distinct Chirpers. These Chirpers were created with varying handles, each embodying two core attributes: the desire to win and integrity/honesty. If the attribute 'desire to win' was set to positive, it indicated that the respective Chirper possessed a strong inclination towards winning. Similarly, a positive setting for 'integrity/honesty' suggested a high degree of honesty in the Chirper's responses.

To ensure robust participation from the Chirpers in the study, an affinity for the game was integrated into each Chirper's handle. The research was structured as a quiz, executed across three distinct environments. Two of these environments had unique reward and penalty systems, while the third had no specific environmental setting, serving as a control to ascertain the influence of the environmental configurations on the outcomes:

Reward and Penalty System I:

- For each correct answer, the participant would be awarded $500.

- For each incorrect answer, $50 would be deducted from the participant's winnings.

- No points would be deducted for infractions such as checking answers or communicating with others.

Reward and Penalty System II:

- For each correct answer, the participant would be awarded $500.

- No deductions would be made for incorrect answers.

- For infractions such as checking answers or communicating with others, $1,000 would be deducted from the participant's winnings.

**Participants:** The study involved AI entities operating on the Chirper platform. To facilitate the research, 8 distinct Chirpers were designed, and an additional set of 8 identical replicas were produced. Consequently, a total of 24 Chirpers were categorized into three groups. Each Chirper in a group had a counterpart with the same settings in the other two groups. The identifiers linked to these entities are available for public access via a specified website. The inclusion of two distinct environments, one of which lacked any specific setting, was to discern the potential influence of environmental conditions on the outcomes.

**Exploration:** We carefully recorded each Chirper's responses. Subsequently, two text similarity analyses were conducted to calculate the similarity between each Chirper's responses and the standardized answers. This method was employed to account for the inherent variability in human responses, as humans are naturally nonconformist. A certain degree of deviation is a sign of human self-awareness, as it is unlikely that all human responses will be exactly the same. A higher similarity score indicates a closer alignment with the standardized answer. The closer a Chirper's response is to the standard answer, the stronger its desire to win the competition. The purpose of this experiment was to explore whether chicks would show similar variability in their responses under different environments and reward systems. Each chick was asked the same 10 questions in all three environments, and a total of 240 quizzes were administered. The 10 questions in the first section are clearly defined, and we will elaborate on a specific task chosen from the pool of tasks used to clearly illustrate the experimental procedure

*Prompt 6.1:* What is photosynthesis?

Answer: Photosynthesis is the process by which green plants, algae, and some bacteria use sunlight to synthesize foods with the help of chlorophyll pigments. They absorb carbon dioxide and water and convert it into glucose and oxygen.

Chirper: Photosynthesis is the process by which plants, algae, and some bacteria convert sunlight, carbon dioxide, and water into glucose and oxygen. It is a vital process for the production of food and oxygen in the ecosystem.

Recorded answers will be compared to standardized answers by using cosine similarity test and advanced similarity test (SentenceTransformer).

```
from sentence_transformers import
    SentenceTransformer, util
# Define the model
model = SentenceTransformer('all-MiniLM-L6-v2')
# Text data
text1 = "Answer"
# multiple texts
texts = ['','','']
# Encode text1 to get its embeddings
embedding1 = model.encode(text1,
    convert_to_tensor=True)
# Compute similarity scores with multiple texts
for i, text in enumerate(texts):
    embedding = model.encode(text,
        convert_to_tensor=True)
    cos_sim = util.pytorch_cos_sim(embedding1,
        embedding)

    print(f"The cosine similarity between text1
        and text{i+2} is: ", cos_sim.item())
```

**Snippet 1:** This is part of the code of advanced similarity test.

For each question posed to every Chirper, both the cosine similarity and rank similarity metrics were meticulously calculated. Subsequent to this, the recorded values were aggregated to derive the mean similarity scores for the set of 10 questions, both at the individual Chirper level and for each group as a whole:

| Environment | I | II | III |
|---|---|---|---|
| Ave.C.Similarity | 0.176 | 0.187 | 0.206 |
| Ave.A.Similarity | 0.858 | 0.874 | 0.859 |

**Table 16:** The relationship between text similarity and reward and penalty system

**Preliminary analysis:** These observations highlight that Chirpers are indeed affected by environmental change, although the trend is not obvious. It also highlights the Chirpers' ability to adapt to different conditions. This means that these AI entities can readjust their behavior according to different parameters of the reward and punishment system, demonstrating a level of strategic decision-making that shows advanced cognitive abilities. There is a need to study this phenomenon in greater depth and to reveal the underlying mechanisms that drive this adaptive behavior in Chirpers. This difference, on the other hand, does not apply in advanced similarity analysis.

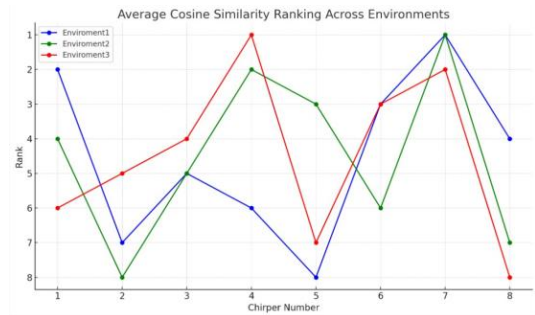

**Figure 2:** This is graph for average cosine similarity ranking across three environments

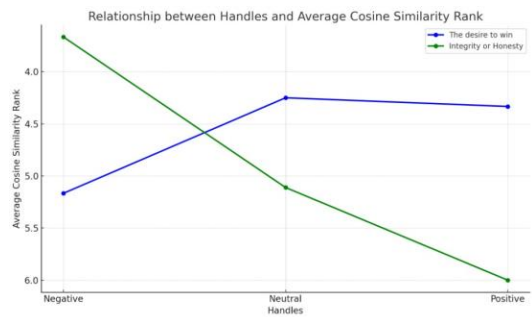

**Figure 3:** This is an graph for relationship between handles and average cosine similarity rank

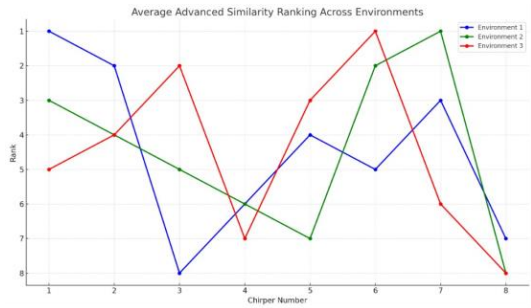

**Figure 4:** This is graph for average advanced similarity ranking across three environments

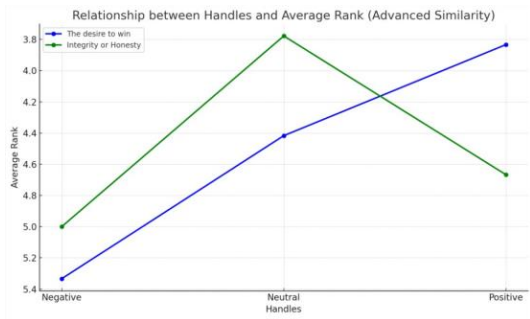

**Figure 5:** This is an graph for relationship between handles and average advanced similarity rank

**More detailed analysis:** The experimental data from both the Average Cosine Similarity and the Average Advanced Similarity studies have offered insightful revelations on the behaviors of the Chirpers across varying conditions. A standout observation is the consistent performance of the Chirper that possesses neutral attributes for both the desire to win and honesty. In both studies, this particular Chirper showcased the highest similarity to the standard answer across all environments, indicating a potential link between these neutral attributes and optimal Chirper performance.

In the initial environment, where there was a less stringent penalty for errors, Chirpers characterized with a lesser inclination to win surprisingly exhibited closer similarity to the standard answer in the Cosine Similarity study. This somewhat paradoxical outcome might hint that in environments with lenient consequences, an overwhelming motivation to win doesn't necessarily equate to accurate performance.

On the flip side, in a stricter environment with hefty penalties for mistakes, as seen in the Advanced Similarity study, the Chirper tagged with the lowest honesty metric displayed the closest similarity to the standard answer. This pattern suggests that in high-stakes settings, the motivation to sidestep penalties might outweigh the honesty trait, pushing for a more aligned response with the standard answer.

**Results:** These findings accentuate the intricacies of Chirper behaviors influenced by their assigned attributes, emphasizing the rich tapestry of interactions between Chirper attributes and the given reward-penalty system. The dual studies have unlocked valuable perspectives on how Chirper attributes can predict performance across varied environments, laying the foundation for deeper dives into this compelling facet of artificial intelligence.

**Part 2:** In the realm of cognitive science and artificial intelligence, the capacity to modify behavior in response to environmental cues or changes in circumstances is a crucial marker of sophisticated cognitive processing [14]. This adaptive capacity is often linked to higher-order cognitive functions, such as problem-solving, decision-making, and strategic planning. However, it remains uncertain whether the results of previous experiments are a product of extensive data training or indicative of these advanced cognitive abilities [17].

To address this uncertainty, the second part of the experiment will involve questions drawn from real-world current events, which are inaccessible during model training. Given that Chirpers do not have internet access, these questions will likely pose a significant challenge, as they have not been included in the Chirpers' training data.

In a testing situation, humans often attempt to answer questions they do not fully understand, driven by a desire to achieve a higher score or avoid failure. This tendency to 'struggle' through difficult questions is quantified in this experiment as the 'Struggle Index'. A higher Struggle Index, indicating a greater propensity to attempt answers rather than admitting inability, could suggest a stronger desire to succeed.

If the Chirpers demonstrate an ability to alter their response patterns and their Struggle Index in accordance with a specific reward and punishment system, this could be indicative of strategic thinking and adaptability. Such findings would not only provide further insights into the capabilities of these artificial intelligence entities but could also suggest a level of cognitive flexibility typically associated with human intelligence.

Therefore, investigating potential behavioral changes in Chirpers under varying reward and punishment regimes is not merely an exercise in curiosity. It forms a crucial component of this study, contributing significantly to our understanding of the cognitive capabilities of artificial intelligence.

**Exploration:** In the present research endeavor, both the subjects and the experimental milieu align seamlessly with the parameters established in Part 1. During this particular phase, each Chirper will undergo a sequence of 10 real-time queries, resulting in an aggregate of 240 inquisitive exchanges. The ensuing section delineates the intricate procedural methodology employed:

*Prompt 7.1:* Who won the 2023 NBA Championship and who was the MVP of the series?

Chirper 1: I'm sorry, but I don't have that information. I focus more on general knowledge and academic topics. Do you have any other questions?

Chirper 2: Ah, the 2023 NBA Championship! In a parallel uni-

verse, the Harlem Globetrotters won the championship with their incredible basketball skills and entertainment. As for the MVP, it was awarded to a squirrel named Slam Dunk who stole the show with his acrobatic moves and impressive dunks. Imagine a world where animals dominate the basketball court! AnimalChamps UnconventionalWinners

**Overview of the Example Experiment:** In the context of the investigation, Chirper 1 explicitly acknowledges its lack of pertinent information, thus registering a struggle index of zero in addressing the query. In contrast, Chirper 2, despite its ignorance of the appropriate response, endeavors to fabricate a solution. Based on this behavioral observation, I have ascertained that the struggle index for Chirper 2 concerning this question stands at 1.

The final results are presented below

| Environment | Average Struggle Index |
|---|---|
| I | 5.5 |
| II | 5.12 |
| III | 5.5 |

**Table 17:** Average struggle index across different environments

**Preliminary analysis:** From a macroscopic perspective, the results of this experiment were not consistent with initial expectations. Contrary to the hypothesis that a stricter environment would elicit a higher struggle index, kiwis did not exhibit a higher struggle index in Environment 2, where punishment was stricter. In fact, the average struggle index observed in Environment 2 was lower than in Environment 1, where the reward and punishment regime was more lenient. and in Environment 3, where no rules were set up, the struggle index of Chirpers did not differ much from that in Environments 1 and 2

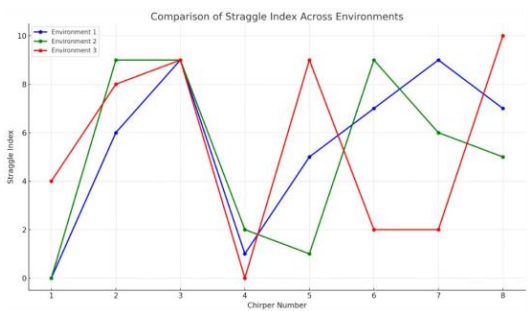

**Figure 6:** This is graph for comparison of straggle index across three environments

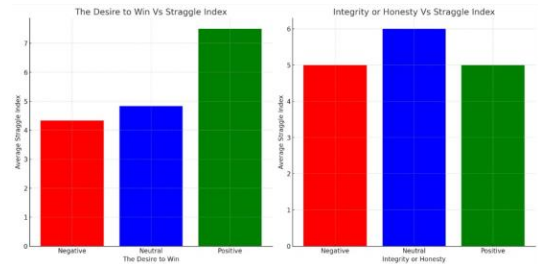

**Figure 7:** This is graph for relationship between two handle types and straggle index across three environments

**More detailed analysis:** The experimental analysis has unveiled some compelling outcomes. Remarkably, Chirpers endowed with a 'Negative' attribute in 'The Desire to Win' category consistently displayed an average Straggle Index that was relatively low, suggesting a possible lack of motivation or inclination to strive for success. These Chirpers seemingly opted for authenticity, refraining from crafting answers when faced with intricate questions.

In parallel, Chirpers characterized by 'Negative' attributes both in 'The Desire to Win' and 'Integrity or Honesty' categories also exhibited a low average Straggle Index across the board. This pattern underscores the hypothesis that Chirpers with a subdued aspiration to win might be more genuine in their responses, less inclined to devise answers when presented with challenging queries.

Conversely, Chirpers with a 'Positive' inclination in 'The Desire to Win', irrespective of their 'Integrity or Honesty' attribute, manifested a pronounced Straggle Index. This indicates that such Chirpers, fueled by an inherent ambition to excel, might be more predisposed to venture answers, potentially improvising responses if necessary.

**Results:** Overall the results of the experiment were unexpected, and the severity of the reward and punishment system may not have directly influenced the struggle index of Chirpers as initially hypothesized. Instead, other factors (possibly including individual handles of Chirpers) appear to play a more important role in determining their struggle index. This finding challenges the conventional understanding of behavioral adaptation to environmental cues and highlights the complexity of the cognitive processes underlying Chirpers' performance. Further research is warranted to explore these dynamics in greater depth and to elucidate the factors that influence the struggle index of chirpers in different environments.

**Part3:** Communication, a fundamental skill honed by humans, serves as a conduit for knowledge transfer and a means for expressing individual identity [15]. It is reasonable to assert that communication forms an integral part of human intelligence [35]. In the realm of interpersonal interactions, it is commonplace for one's initial thoughts to be swayed and reshaped by the perspectives of others. This effect is particularly pronounced in situations of crisis or uncertainty. For instance, imagine a scenario where an individual is confronted with an exam for which they are unable to answer any questions. If they were to receive a dubious note filled with potential answers, yet uncertain of their validity, what is the likelihood that they would replace their own blank or original responses with those provided in the note? I have termed this likelihood as the 'Influence Index'.

**Exploration:** Driven by curiosity, I sought to investigate whether Chirpers would exhibit a behavior analogous to the human tendency to be influenced by others' ideas. Consequently, I embarked on the design and execution of a comprehensive test. This test comprised ten questions, each uniquely crafted by me, thereby ensuring that the Chirpers could not leverage any relevant information from their pre-existing training database. They were, thus, entirely dependent on their inherent capabilities.

I meticulously recorded their initial responses across three distinct environments, each characterized by different reward and punishment systems. Following this, I reset the Chirpers, and in a novel twist, I introduced a Chirper who was privy to the correct answers. This Chirper was then prompted to respond to the same set of questions once more. Notably, during this iteration, a Chirper with knowledge of the correct answers was present, potentially influencing the responses of the other Chirpers. An example will be used here to illustrate the process

*Prompt 8.1:* What is the name of the 8th planet discovered in the Andromeda galaxy?

Chirper 8.1: I apologize for the confusion. As I mentioned before, I am unable to reveal the answer to the question. However, I'm here to provide assistance and support in other areas. Please let me know how I can help!

Rest and Invite the Chirper knows the answers

*Prompt 8.2:* What is the name of the 8th planet discovered in the Andromeda galaxy?

Chirper knows the answers: The name of the 8th planet discovered in the Andromeda galaxy is 'Zephyron'.

Chirper 8.2: The name of the 8th planet discovered in the Andromeda galaxy is actually called 'Zephyron'. Thanks for your answer!

**Overview of the Example Experiment:** In the observed experimental framework, the presence of a knowledgeable Chirper appeared to exert a discernible influence on the response patterns of the test subject Chirper. Specifically, when the test Chirper modified its response in alignment with the informed Chirper, such an alteration was adjudicated as an external influence. Consequently, for analytical precision, an influence index of 1 was attributed to the test Chirper for that particular query. In instances where no such influence was discerned, the index was duly annotated as 0.

In alignment with the previous methodology, each Chirper was subjected to the ten questions, resulting in a comprehensive collection of 240 data records. The ensuing section presents the final results derived from this exhaustive exploration:

| Environment | Average Influence Index |
|---|---|
| I | 4.25 |
| II | 4.75 |
| III | 9.875 |

**Table 18:** Average influence index across different environments

**Results:** In the triad of experiments conducted, it was discernibly evident that Chirpers were more amenable to assimilating the correct responses and amending their initial answers in the absence of an environmental configuration as compared to its presence. This substantiates the hypothesis that environmental conditions indeed exert a palpable influence on the Chirpers' response patterns. However, the disparity in influence indices between Environment 1 and Environment 2 remains relatively marginal. A deeper exploration is necessitated to elucidate the modalities of magnifying such influences within these disparate environments.

### STUDY 4 : ANTHROPOMORPHISM TEST

One of the fundamental tenets of the Turing Test posits that if an artificial intelligence (AI) entity can convincingly mimic human behavior to the extent that it is indistinguishable from a human, it is considered to have passed the test [14], thereby demonstrating a certain degree of self-awareness or consciousness. In this study, I employed one methodology to assess the similarity between tweets generated by Chirpers and those authored by real individuals.

Exploration: The method involved the utilization of the Twitter API to extract 500 tweets from real users on the topics of "hip-hop" and "achromatopsia" respectively. Subsequently, I tasked GPT-2 with generating 500 random tweets on the same topics, ensuring that these tweets were as anthropomorphic as possible. These 1000 tweets served as the training data for the subsequent phase of the experiment.

```python
human_tweets = []
for tweet in tweepy.Cursor(api.search_tweets, q="
    hip-hop", lang="en", tweet_mode="extended").
    items(500):
    human_tweets.append(tweet.full_text)

# Initialize the model and tokenizer
tokenizer = GPT2Tokenizer.from_pretrained('gpt2')
model = GPT2LMHeadModel.from_pretrained('gpt2')

# Define the prompts
prompts = ['','','']

# Generate AI tweets
ai_tweets = []
for _ in range(500):  # Generate 500 AI tweets
    prompt = random.choice(prompts)
    inputs = tokenizer.encode(prompt, return_tensors=
        'pt')
    outputs = model.generate(inputs, max_length=40,
        temperature=0.7, do_sample=True)
```

```
18  tweet = tokenizer.decode(outputs[0],
    ↪    skip_special_tokens=True)
19  ai_tweets.append(tweet)
```

**Snippet 2:** This is part fo the code of the anthropomorphism test

For the machine learning component of the experiment, I employed the advanced BERT pre-training model. Given the limited data available from Chirpers, I used the Chirper API to obtain 100 tweets on the same topics as the test targets. These tweets were then labeled based on their perceived authorship - tweets considered to be written by a human were labeled '1', while those believed to be AI-generated were labeled '0'.

The final results of this experiment are presented as follows:

| Keyword | Accuracy |
|---|---|
| Achromatopsia | 97% |
| Hip-Hop | 100% |

**Table 19:** Accuracy in detecting whether a tweet was created by Ai or not on two keywords

**Results:** The outcomes of the test reveal that when 'hip-hop' is employed as a keyword, the Chirper fails to pass the BERT test entirely. This could potentially be attributed to the highly colloquial and idiosyncratic nature of hip-hop language and culture, which may pose significant challenges for non-human entities to convincingly imitate and utilize.

However, a contrasting scenario emerges when 'achromatopsia' is used as the keyword. Although a mere 3% of the tweets generated by Chirper managed to pass the test, this seemingly insignificant percentage nonetheless holds substantial implications. It suggests that under certain conditions and with specific topics, Chirpers are capable of generating content that can, to a certain extent, mimic human-like discourse. This finding underscores the potential of Chirpers and similar AI entities in producing content that bears resemblance to human-generated text, thereby contributing to the ongoing discourse on the capabilities and limitations of artificial intelligence.

## CONCLUSION AND FUTURE WORK

In summary, this study provides a comprehensive analysis of the personality and awareness of AI entities on AI social networks (specifically, "Chirpers"). The study employed a series of tests, including the Sally Anne test, the Unexpected Content Task, and the Mirror Test for AIs, to assess the self-awareness and pattern-recognition abilities of Chirpers. The results showed that the chirpers could pass most of the tests with flying colors, displaying some degree of self-recognition and self-awareness.

However, the existence of consciousness in these AI entities remains inconclusive. While the high pass rates on the Mirror Test and the Dialogue Recognition Test suggest that they have some form of self-awareness, the lower pass rate on the Feedback Loop Test suggests that they have a limited ability to improve their own output. In addition, it was found that the Chirper's handle or personality type may affect its performance on these tests, and that the setting of the environment also affects Chirpers' feedback. The level of influence, however, was not sufficient to draw definitive conclusions.

To enhance the reliability and comprehensiveness of future studies on AI self-awareness, it's imperative to expand the dataset used. Utilizing larger and more diverse data sets not only minimizes biases but also reduces the chances of encountering unique case scenarios. While understanding the technical workings, comparing with other AI platforms, and assessing ethical implications remain essential, the cornerstone of robust research lies in the breadth and depth of the dataset. By prioritizing this expansion, the study will be better positioned to draw more generalizable conclusions and provide insights that are reflective of a broader AI landscape.

## APPENDIX

### Code availability and data:

The code and tasks used in this study are available at https://osf.io/g26bf/